\documentclass[sigconf]{acmart}

\AtBeginDocument{%
  }

\copyrightyear{2024}
\acmYear{2024}
\setcopyright{acmlicensed}\acmConference[NLBSE '24]{2024 ACM/IEEE International Workshop on NL-based Software Engineering}{April 20, 2024}{Lisbon, Portugal}
\acmBooktitle{2024 ACM/IEEE International Workshop on NL-based Software Engineering (NLBSE '24), April 20, 2024, Lisbon, Portugal}
\acmDOI{10.1145/3643787.3648044}
\acmISBN{979-8-4007-0576-2/24/04}

\usepackage{graphicx} 
\usepackage{adjustbox}
\usepackage{listings}
\usepackage{natbib}
\usepackage{enumitem}
\usepackage{multirow}
\usepackage{pifont}
\newcommand{\cmark}{\ding{51}}%
\newcommand{\xmark}{\ding{55}}%

\begin{document}

\title{Dopamin: Transformer-based Comment Classifiers through Domain Post-Training and Multi-level Layer Aggregation}

\author{Nam Le Hai}
\email{namlh35@fpt.com}
\affiliation{%
	\institution{FPT Software AI Center}
	\city{Hanoi}
	\country{Vietnam}
}

\author{Nghi D. Q. Bui}
\email{nghi.bui@fulbright.edu.vn}
\affiliation{%
	\institution{Fulbright University}
	\city{Ho Chi Minh City}
	\country{Vietnam}
}

\begin{abstract}
Code comments provide important information for understanding the source code. They can help developers understand the overall purpose of a function or class, as well as identify bugs and technical debt. However, an overabundance of comments is meaningless and counterproductive. As a result, it is critical to automatically filter out these comments for specific purposes. In this paper, we present \textbf{Dopamin}, a Transformer-based tool for dealing with this issue. Our model excels not only in presenting knowledge sharing of common categories across multiple languages, but also in achieving robust performance in comment classification by improving comment representation. As a result, it outperforms the STACC baseline by 3\% on the NLBSE'24 Tool Competition dataset in terms of average F1-score, while maintaining a comparable inference time for practical use. The source code is publicity available at \url{https://github.com/FSoft-AI4Code/Dopamin}.
	
\end{abstract}

\maketitle

\section{Introduction}
In the intricate world of software development, source code comments play a crucial role, serving as the backbone of applications by elucidating the functionality and intent behind code segments. These comments vary widely in their utility, ranging from summarizing the purpose of functions or classes, aiding in code maintenance, to identifying instances of technical debt, as highlighted in studies like \citep{manh2023vault, li2023automatic, DBLP:conf/acl/ToBGN23}. However, not all code comments are equally beneficial, or different information in comments can be used for different development tasks. With the increasing complexity of software projects, the task of discerning valuable comments from the less pertinent ones has become more critical.

Current tools for comment classification, as discussed in references like \cite{pascarella2017, rani2021, AlKaswan2023}, predominantly focus on categorizing comments based on their apparent utility in the coding workflow. Despite their effectiveness, these tools often lack the nuanced understanding needed to differentiate between subtly different types of comments, limiting their practical utility. Recognizing this gap, this paper introduces \textbf{Dopamin} - a Transformer-based Comment Classifier that utilizes a code language model for an enhanced classification process.

Dopamin takes a novel approach, relying on domain post-training on diverse comment types across various coding environments, and adopting a multi-level layer aggregation strategy inspired by \citet{DBLP:conf/icnlsp/KarimiRP21}. The post-training procedure incorporates data from all programming languages, facilitating knowledge transfer across different languages and leveraging the high-resource language (Java) to improve the less resource-intensive languages (Python). Meanwhile, layer aggregation methodology enables Dopamin to not only classify comments but also understand the nuanced semantic information they carry, as the higher layers in BERT are adept at capturing intricate semantic features, a concept supported by \citet{jawahar2019does}. By doing so, Dopamin significantly improves the relevance and accuracy of comment classification, catering to the evolving complexity of software development.

The efficacy of Dopamin is evident in our experimental results, where it achieves an F1-score of 0.74, surpassing the 0.71 F1-score of the existing STACC \cite{AlKaswan2023} baseline. Through Dopamin, we aim to redefine the standards in code comment classification, providing a tool that is effective in distinguishing various types of comments on multiple programming languages.

\begin{table}
\caption{NLBSE’24 Tool Competition dataset properties}
\label{tab:dataset}
\begin{adjustbox}{width=0.45\textwidth}
    \begin{tabular}{lcc}
        \toprule
        Language & Number of categories & Number of data per category\\
        \midrule
        Java & 7 & 10.555 \\
        Python & 5 & 2.555 \\
        Pharo & 7 & 1.765 \\
        \bottomrule
    \end{tabular}
\end{adjustbox}
\end{table}

\section{Data preparation}

In this section, we illustrate the NLBSE’24 Tool Competition dataset introduced by \citet{nlbse2024}, detailing our approach to processing comments and splitting the training data for model selection.

\subsection{Dataset statistic}

The competition provided binary comment classification data of three languages (Python, Java, and Pharo). In total, there are 19 categories corresponding to 19 classifiers required to build. Overall information of the dataset is shown in Table  \ref{tab:dataset} and details on the competition repository\footnote{\url{https://github.com/nlbse2024/code-comment-classification}}.

\subsection{Data preprocess}
\label{sec:data_preprocess}
\textbf{Input feature}: Follow \citet{AlKaswan2023},   we concatenate the class name and comment sentence to serve as the input to the model, employing "</s>" as the separator between them. \\
\textbf{Data spliting}: The training data is divided into a training set and a validation set for refining the optimal model. Instead of employing the validation set for hyperparameter tuning, it is utilized to choose the best checkpoint during the training process. Due to the modest amount of data, we select only 10\% of the training set of each category to serve as the validation set. We employ stratified sampling to maintain the label distribution in both sets. 

\begin{table}
\caption{Different model results on Validation set}
\label{tab:model_selection}
\begin{adjustbox}{width=0.35\textwidth}
\scriptsize
\begin{tabular}{lccc}
\toprule
Model & Precision & Recall & F1\\
\midrule
CodeBERT & 0.7921 & 0.8438 & \textbf{0.8133} \\
RoBERTa & 0.7900 & 0.8304 & 0.8063 \\
ALBERT & 0.6695 & 0.7769 & 0.7108 \\
\bottomrule
\end{tabular}
\end{adjustbox}
\end{table}

\section{Methodology}

This section describes the Dopamin methodology, including model selection, the methodology for obtaining the optimal checkpoint, domain post-training procedures, and layer aggregation techniques.

\subsection{Model selection}
We investigate several candidates as the backbone model. Since the primary source of classification information comes from comments, which are in natural language form, we choose RoBERTa \cite{liu2019roberta} and ALBERT \cite{lan2019albert} as candidates. Additionally, the comments primarily contain syntax related to the coding domain. Therefore, we are also considering CodeBERT \cite{feng2020codebert}, which is a language model pretrained on large code corpus, as an option. We opt for these architectures because they are based on Transformer encoders, commonly employed for classification tasks. Additionally, the base versions of these models share the same size as the baseline (STACC), resulting in fair comparison and no additional overhead in inference time, which is a key consideration for evaluation score. The performance of each model on the validation set is presented in Table \ref{tab:model_selection}. As a result, we select CodeBERT - the model that achieves the best F1 score on the validation set, as the backbone model.

\subsection{Domain post-training}
Table \ref{tab:dataset} shows that Python and Pharo have much fewer examples compared to Java. Besides, some categories are contained in both Java and Python languages such as \textit{Expand}, \textit{Summary}, and \textit{Usage}. Therefore, we combined the data of all languages to finetune the CodeBERT backbone model in the data domain (post-training), facilitating the transfer of knowledge across languages before individual training of models for each category in the target domain.

During the post-training procedure, we did not concatenate the class name and
comment sentence as input to the model as mentioned in Section \ref{sec:data_preprocess}. Instead, we concatenate the category to the comment sentence since the model is required to predict for all categories, necessitating the inclusion of category information in the input.

Following the post-training of the model in the target domain, we utilize it as the initial state for the model to be trained individually for each category using the procedure in Section \ref{sec:optimal_ckpt} and input feature in Section \ref{sec:data_preprocess}.

\subsection{Multi-level layer aggregation}
\begin{figure}[t]
\includegraphics[width=0.48\textwidth]{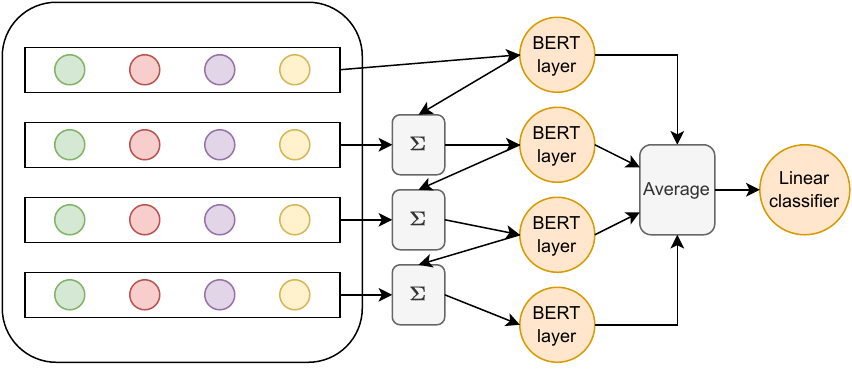}
\centering
\caption{Hierarchical aggregation}
\label{fig:hsum}
\end{figure}

\citet{jawahar2019does} previously showed that the upper layers in BERT yield rich semantic features of linguistic information and distinct layers can exhibit distinct capabilities in encoding semantic information. Hence, combining these layers can obtain a more comprehensive representation for the input text. Therefore, we adopt the Hierarchical aggregation (HSUM) introduced by \citet{DBLP:conf/icnlsp/KarimiRP21} to enrich the comment representation. The illustration of HSUM is shown in Figure \ref{fig:hsum}. Specifically, we combined the top four layers of the model to obtain the final representation for comment.

\subsection{Optimal checkpoint}
\label{sec:optimal_ckpt}
Given the high cost of hyperparameter search for 19 categories, we choose to \textbf{keep the hyperparameters constant} throughout the training process for each category. Instead, we use the validation set to determine the best checkpoint step. This strategy is similar to early stopping, which aims to prevent the model from overfitting. We employ it as a heuristic to determine the optimal step for the stage of training the model on the original training set without validation. Specifically, there are two stages in the training process.

\begin{itemize}[leftmargin=*]
\item Stage 1: the model is trained on the training set and the best checkpoint ($optimal\_step$) is obtained based on the F1-score of the validation set.
\item Stage 2: Considering the limited amount of data, training the model on the entire original training set is essential. After obtaining the $optimal\_step$ in stage 1, we train the model on the original training set and acquire the final model at step $optimal\_step + extra\_steps$. The $extra\_steps$ represents the additional steps due to the incorporation of more data during training.
\end{itemize}

For example,  after \textit{stage 1} the model for the \textit{Java - Deprecation} category attains its highest F1 score at step 200. The $extra\_steps$ is set to 100, thus after training on the full original dataset in \textit{stage 2}, we obtain the checkpoint at step 300 as the final model for the \textit{Deprecation} category of Java.




\section{Experiment setup}
\subsection{Training hyperparameters}

For reproducibility, we set the random seed as 0. 
The hyperparameters are selected based on references from prior studies \citep{feng2020codebert,liu2019roberta, lan2019albert} that involved fine-tuning models on downstream tasks.

\begin{itemize}[leftmargin=*]
\item \textbf{Post-training stage}: In this stage, we train the backbone model during 10 epochs with the learning rate of $2e-5$, batch size of 64, and evaluation step of 500.

\item \textbf{Invidual classifier training stage}: Each model is trained for 10 epochs for Java categories and 20 epochs for Python or Pharo categories. 
We assign a lower number of epochs to Java due to its higher volume of training examples, resulting in a greater number of training steps within a single epoch. The learning rate is set to 1e-5, the batch size is 64, and the evaluation step is 50. We use the $extra\_steps$ of 100 for all categories.
\end{itemize}

\subsection{Implementation}
We use the HuggingFace transformers\footnote{\url{https://huggingface.co/docs/transformers/index}} and PyTorch packages\footnote{\url{https://pytorch.org/}} to implement Dopamin. All experiments are conducted using two Nvidia A100 GPUs with 80GB of VRAM. During the evaluation on the provided test set, we utilize Google Colab T4, adhering to the competition's specifications, to acquire the inference time.

\subsection{Metrics}
For evaluation, we employ the metrics outlined by the competition, considering a category $c$. Specifically, we calculate the recall $R_c$, precision $P_c$, and F1 score $F1_c$ for each category. These metrics are defined as follows:

\begin{equation}
P_c=\frac{TP_c}{TP_c + FP_c}, R_c=\frac{TP_c}{TP_c + FN_c}, F1_c=2.\frac{P_c.R_c}{P_c + R_c} \nonumber  
\end{equation}

in which, $TP_c$, $FP_c$, and $FN_c$ are the true positives, false positives, and false negatives for a category $c$, correspondingly.

Finally, the submission score of the competition using both the average F1-score and the inference time is defined as:
\begin{align}
	& submission\_score \nonumber \\
	& = 0.75(avg. F1) + 0.25\frac{max\_avg\_runtime - measured\_avg\_runtime}{max\_avg\_runtime} \nonumber
\end{align}

\section{Experimental results}

In this section, we present the performance comparison of Dopamin against the STACC baseline. Table~\ref{tab:performance} shows the performance comparison between Dopamin and STACC across various categories and languages demonstrates Dopamin's enhanced capabilities in comment classification. Below is a summary of the key findings from the table.

\subsubsection{Overall Performance}
Dopamin attains a comprehensive F1-score of \textbf{0.74}, balancing Precision at 0.73 and Recall at 0.75. This marks an enhancement compared to STACC's overall F1-score of 0.71. Moreover, Dopamin incurs no additional inference time overhead compared to STACC. Consequently, the submission score reaches \textbf{0.703} compare to 0.675 of STACC baseline. Overall, the result indicates a consistent and significant enhancement in classification effectiveness.

\subsubsection{Performance by Language}

\begin{itemize}[leftmargin=*]
\item \textbf{Java}: Dopamin performs better in categories like \textit{Pointer}, \textit{Summary}, \textit{Ownership}, \textit{Rational}, and \textit{Usage}, with notable improvements in F1-scores. For instance, Dopamin achieves F1-score at 0.90 compared to STACC's 0.78 in the \textit{Summary} category.

\item \textbf{Pharo}: Dopamin shows improved performance in categories like \textit{Classreferences}, \textit{Collaborators}, and \textit{Example}. For example, in the \textit{Classreferences} category, Dopamin's F1-score is 0.68 compared to STACC's 0.52. However, Dopamin exhibits a weakness in 4 out of 7 categories when compared to STACC in this language. The explanation might stem from Pharo being the language with the lowest data volume per category (under 2000). Given the limited data, the few-shot approach (STACC) appears more effective than fine-tuning with classification loss (Dopamin). Moreover, the absence of overlap categories between Pharo and other languages could restrict the advantages of the Post-training stage.

\item \textbf{Python}: Dopamin also outperforms STACC in the Python category, particularly in \textit{Parameters}, \textit{Summary}, and \textit{Usage} categories. These categories overlap with those in Java, suggesting the effectiveness of knowledge transfer during our Post-training process.
\end{itemize}

\begin{table}
\caption{Ablation study on our proposed components: Domain Post-training (PoT) and Layer aggregation (LA). All the models use CodeBERT as the backbone.}
\label{tab:ablation}
\scriptsize
\begin{adjustbox}{width=0.48\textwidth}
\begin{tabular}{ll|ccc|ccc}
\toprule
\multirow{2}*{PoT} & \multirow{2}*{LA} & \multicolumn{3}{c|}{Validation set} & \multicolumn{3}{c}{Test set} \\
\cmidrule{3-8}
 & & Precision & Recall & F1 & Precision & Recall & F1\\
\midrule
\cmark & \cmark & 0.844 & 0.905 & \textbf{0.869} & 0.735 & 0.745 & \textbf{0.738}\\
\cmark & \xmark & 0.851 & 0.892 & 0.866 & 0.712 & 0.729 & 0.717 \\
\xmark & \cmark & 0.820 & 0.824 & 0.820 & 0.727 & 0.720 & 0.719\\
\xmark & \xmark & 0.792 & 0.844 & 0.813 & 0.728 & 0.713 & 0.714 \\
\bottomrule
\end{tabular}
\end{adjustbox}
\end{table}

\begin{table*}
\caption{Performance of Dopamin against the STACC baseline}
\label{tab:performance}
\begin{adjustbox}{width=0.66\textwidth}
\scriptsize
\begin{tabular}{ll|ccc|ccc|c}
\toprule
\multirow{2}*{Language} & \multirow{2}*{Category} & \multicolumn{3}{c|}{STACC} & \multicolumn{3}{c|}{Dopamin} & \multirow{2}*{$\bigtriangleup F1_c$} \\
\cmidrule{3-8}
 &  & $P_c$  & $R_c$ & $F1_c$ & $P_c$ & $R_c$ & $F1_c$ & \\
\midrule
Java & Deprecation & 0.81 & 0.94 & 0.87 & 0.81 & 0.88 & 0.85 & -0.02\\
Java & Pointer & 0.82 & 0.78 & 0.80 & 0.89 & 0.81 & 0.85 & +0.05\\
Java & Summary & 0.93 & 0.67 & 0.78 & 0.94 & 0.87 & 0.90 & +0.12\\
Java & Expand & 0.57 & 0.73 & 0.64 & 0.52 & 0.63 & 0.57 & -0.07 \\
Java & Ownership & 0.97 & 1.0 & 0.99 & 1.0 & 0.99 & 1.0 & +0.01\\
Java & Rational & 0.49 & 0.51 & 0.5 & 0.49 & 0.59 & 0.54 & +0.04\\
Java & Usage & 0.64 & 0.92 & 0.76 & 0.87 & 0.95 & 0.91 & +0.15\\
\midrule
Pharo & Classreferences & 0.47 & 0.57 & 0.52 & 0.76 & 0.62 & 0.68 & +0.16\\
Pharo & Example & 0.93 & 0.89 & 0.91 & 0.93 & 0.93 & 0.93 & +0.02\\
Pharo & Keyimplementationpoints & 0.69 & 0.79 & 0.73 & 0.5 & 0.69 & 0.58 & -0.15\\
Pharo & Collaborators & 0.36 & 0.91 & 0.51 & 0.57 & 0.64 & 0.60 & +0.09\\
Pharo & Intent & 0.87 & 0.89 & 0.88 & 0.87 & 0.85 & 0.86 & -0.02\\
Pharo & Keymessages	& 0.79 & 0.91 & 0.85 & 0.86 & 0.81 & 0.83 & -0.02\\
Pharo & Responsibilities & 0.67 & 0.63 & 0.65 & 0.59 & 0.62 & 0.61 & -0.04\\
\midrule
Python & Developmentnotes & 0.43 & 0.54 & 0.48 & 0.45 & 0.44 & 0.44 & -0.04 \\
Python & Parameters & 0.78 & 0.86 & 0.81 & 0.88 & 0.85 & 0.86 & +0.05\\
Python & Summary & 0.62 & 0.64 & 0.63 & 0.8 & 0.69 & 0.74 & +0.11\\
Python & Expand & 0.52 & 0.56 & 0.54 & 0.51 & 0.53 & 0.52 & -0.02\\
Python & Usage & 0.69 & 0.77 & 0.73 & 0.72 & 0.79 & 0.76 & +0.03\\
\midrule
Overall & & 0.69 & 0.76 & 0.71 & 0.73 & 0.75 & \textbf{0.74} & +0.03\\
\bottomrule
\end{tabular}
\end{adjustbox}
\end{table*}

\subsubsection{Category-Specific Performance}

Dopamin excels particularly in categories where understanding the context and semantics is crucial, like \textit{Summary}, \textit{Usage}, and \textit{Ownership}.
In some categories, like \textit{Java - Expand} and \textit{Pharo - Keyimplementationpoints}, Dopamin's performance is lower than STACC. This suggests room for further optimization in certain specific categories.

In summary, Dopamin generally exhibits a balanced improvement in both precision ($P_c$) and recall ($R_c$) across various categories. For example, in the \textit{Java - Usage} category, Dopamin shows a precision of 0.87 and recall of 0.95, compared to STACC's 0.64 and 0.92, respectively.
Dopamin also demonstrates a notable improvement over STACC in most categories and languages, reflecting its advanced capability in understanding and classifying code comments.

\section{Ablation study}
In this section, we present the experimental outcomes obtained by systematically trimming individual components within Dopamin, showcasing the efficacy of each element.  The results of this study on Validation and Test sets are illustrated in Table \ref{tab:ablation}.

On the validation set, improvements are evident when adopting Post-training or Layer Aggregation independently, as opposed to fine-tuning the unmodified CodeBERT model. Especially, the post-training method makes up significant enhancement (over 5\% on F1 score). Meanwhile, on the test set, HSUM demonstrates its efficacy in enhancing F1 score performance. This underscores the effectiveness of each proposed component. Consequently, the combination of these two methodologies, encapsulated in Dopamin, attains the highest performance of F1 score on both evaluation sets.

\section{Conclusion}
We proposed Dopamin, a novel approach for code comment classification, which demonstrates notable advancements in the benchmark. By selecting CodeBERT as the backbone model and implementing an optimal checkpoint process, we have optimized the classification accuracy across various programming languages. The domain post-training procedure significantly enhances performance for low-resource languages, illustrating the model's adaptability. Additionally, the use of multi-level layer aggregation, specifically the Hierarchical aggregation (HSUM) technique, enriches the semantic representation of comments, contributing to Dopamin's superior performance over the STACC baseline in most categories. Overall, Dopamin's methodology and results mark a significant improvement in the efficiency and precision of code comment classification, offering valuable insights and tools for software communities.

\bibliographystyle{dopamin}
\bibliography{dopamin}

\end{document}